\documentclass{article}

\usepackage[final,nonatbib]{nips_2017}

\usepackage[utf8]{inputenc} 
\usepackage[T1]{fontenc}    
\usepackage{hyperref}       
\usepackage{url}            
\usepackage{booktabs}       
\usepackage{amsfonts}       
\usepackage{nicefrac}       
\usepackage{microtype}      
\usepackage{algorithm}
\usepackage{algorithmic}
\usepackage{graphicx}
\usepackage[normalem]{ulem}
\usepackage{amsmath}
\usepackage{amssymb}
\DeclareMathOperator*{\argmin}{argmin}
\usepackage{amsthm}
\theoremstyle{plain}
\newtheorem{thm}{Theorem}
\theoremstyle{definition}
\newtheorem{defn}[thm]{Definition}
\usepackage{makecell}
\newcommand{\tabincell}[2]{\begin{tabular}{@{}#1@{}}#2\end{tabular}}
\usepackage{mathtools}
\newcommand\iidsim{\stackrel{\mathclap{\normalfont\mbox{i.i.d.}}}{\thicksim}}
\usepackage{caption}
\title{Optimal Transport for Deep Joint Transfer Learning}

\author{
  Ying Lu \qquad Liming Chen \qquad Alexandre Saidi \\
  Ecole Centrale de Lyon, France\\
  \texttt{\{ying.lu, liming.chen, alexandre.saidi\}@ec-lyon.fr} \\
}

\begin{document}

\maketitle

\begin{abstract} 
  Training a Deep Neural Network (DNN) from scratch requires a large amount of labeled data. For a classification task where only small amount of training data is available, a common solution is to perform fine-tuning on a DNN which is pre-trained with related source data. This consecutive training process is time consuming and does not consider explicitly the relatedness  between different source and  target tasks.
  In this paper, we propose a novel method to jointly fine-tune a Deep Neural Network with source data and target data. By adding an Optimal Transport loss (OT loss) between source and target classifier predictions as a constraint on the source classifier, the proposed Joint Transfer Learning Network (JTLN) can effectively learn useful knowledge for target classification from source data. Furthermore, by using different kind of metric as cost matrix for the OT loss,  JTLN can incorporate different prior knowledge about the relatedness between target categories and source categories.
  We carried out experiments with JTLN based on Alexnet on image classification datasets and the results verify the effectiveness of the proposed JTLN in comparison with standard consecutive fine-tuning. This Joint Transfer Learning with OT loss is general and can also be applied to other kind of Neural Networks. 
\end{abstract}

\section{Introduction}
\label{sec:intro}
  
    Supervised machine learning generally requires a large amount of labeled training data for an effective training of the underlying prediction model, especially when the prediction model is complex, \textit{e.g.}, Deep Neural Networks (DNN), where the number of parameters is at a scale of thousand millions. However in practice, collecting a sufficient number of manually labeled training samples may prove tedious, time consuming, even impractical,  and therefore prohibitive, \textit{e.g.}, object edge detection, medical image segmentation, where a pixel-wise ground truth is needed. This is all the more true when the task is novel or rare. For example, for a fine-grained image classification task, for some rare categories we can only gather very limited number of image samples. Transfer Learning (TL) aims to leverage existing related source domain data  for an informed knowledge transfer to a target task and thereby solve or mitigate this kind of "data starvation" problem  to help the learning of a target task. As such, TL has received an increasing interest from several research communities \cite{SurveyTL} \cite{shao2015transfer}.
 
  In this paper we consider the Inductive Transfer Learning (ITL)
  problem \cite{SurveyTL}, which aims at learning an effective classification model for some target categories with few training samples, by leveraging knowledge from different but related source categories with far more training samples. Given the breakthrough of Deep Neural Networks (DNNs) in an increasing number of applications, a natural yet simple  solution to this problem consists of fine-tuning a DNN which is pre-learned on some related source data for a given target classification task \cite{yosinski2014transferable}. However, although this fine-tuning process can inherit or preserve the knowledge learned during pre-training on source data, prior knowledge about the relatedness between source  and target tasks is not explicitly explored. As a result, such a fine-tuning process may fall short to achieve an effective adaptation of a pre-trained DNN for a given target task , especially  when the latter has very few labeled data.  
  
  A recent move is \cite{ge2017borrowing} on selective joint fine-tuning, which tackles this problem by first selecting relevant samples in the source domain, then performing a joint fine-tuning on target training data and the selected source data. Although the selective step ensures the fine-tuning process to use only source samples which are related to a given target domain, in the joint fine-tuning step the source classifier and target classifier are still trained as two different classifiers. 

 In this paper, we propose to explicitly account for the relatedness between source and target tasks and explore such prior knowledge through the design of a novel loss function, namely Optimal Transport loss (OT loss), which is minimized during joint training of the underlying neural network, in order to bridge the gap between the source and target classifiers.  This results in a Joint Transfer Learning Network (JTLN). This JTLN can be built upon common Deep Neural Network structure. 
In JTLN, the source data and target data go through same feature extraction layers simultaneously, and then separate into two different classification layers. The Optimal Transport loss is added between the two classification layers' outputs, in order to minimize the distance between two classifiers' predictions. 
  
  
  As the Optimal Transport loss is calculated with a pre-defined cost matrix, this JTLN can therefore incorporate different prior knowledge about the relations between source and  target  tasks by using different kinds of cost metrics. In this work, we show two examples of using the distance between category distributions as cost metric. 
  
  
  The contributions of this paper are threefold: 
  \begin{enumerate}
  \item We propose a Joint Transfer Learning framework built upon existing Deep Neural Networks for Inductive Transfer Learning.
  \item We extend the Wasserstein loss proposed in \cite{frogner2015learning} to a more general Optimal Transport loss for comparing probability measures with different length, and use it as a soft penalty in our JTLN.
  \item We show two different ways of using the distance between category distributions as cost metric for OT loss. Experimental results on two ITL image classification datasets show that JTLN with these two cost metrics can achieve better performance than consecutive fine-tuning or simple joint fine-tuning without extra constraint.
  \end{enumerate}

\section{Related work}
\label{sec:related_work}

A related problem in transfer learning (TL) is the Domain Adaptation (DA) problem, which is Transductive TL \cite{SurveyTL} and assumes that the source domain and target domain share the same label space, while following different probability distributions. Optimal Transport has already been successfully applied to DA in \cite{courty2016optimal}. Recently several deep joint learning methods have been proposed to solve this problem. For example in \cite{long2015learning} the authors propose to add multiple adaptation layers upon deep convolutional networks. Through these adaptation layers the mean embeddings of source distribution and target distribution are matched, therefore encouraging the network to learn a shared feature space for source domain and target domain. In \cite{long2016unsupervised} the authors extend the previous work by adding additional residual layers to adjust classifier mismatch between source domain and target domain. Although these methods work well for domain adaptation problems, their assumption that the source domain and target domain share a same label space and have a limited distribution discrepancy restrict the possibility of applying these methods for Inductive Transfer Learning. 

Until recently most state-of-the-art ITL methods are based on shallow machine learning models \cite{shao2015transfer} \cite{kuzborskij2013from} \cite{kuzborskij2015transfer} \cite{li2017self}. For example in \cite{kuzborskij2015transfer} the authors propose to select relevant source hypotheses and feature dimensions through greedy subset selection. In \cite{li2017self} the authors propose to learn a high quality dictionary for low-rank coding across source domain and target domain for self-taught learning (which is ITL with only unlabeled samples in source domain). To the best of our knowledge, the proposed JTLN is the first work to tackle ITL with Deep Neural Networks and optimal transport theory.

The proposed JTLN has been inspired by a recent work on wasserstein loss \cite{frogner2015learning}. Frogner et al. proposed a wasserstein loss as a soft penalty for multi-label prediction. Although their wasserstein loss is calculated between predicted label vector and ground-truth label vector with the same length, the matrix scaling process used to calculate this wasserstein loss is actually not restricted to square transportation matrix. In this paper, we extend this wassertein loss to a more general Optimal Transport loss for label vectors with different length. As a result,  the proposed JTLN enables the exploration of prior knowledge through the initial cost matrix and makes use of  the OT loss as a soft penalty for bridging the gap between target and source classifier predictions.


  
\section{Joint Transfer Learning Network}
\label{sec:JTLN}

\subsection{Problem definition and the JTLN structure}
\label{subsec:problem_definition}

Assume that we have a small target training set $\mathcal{T}= \{ (\mathbf{x}^{t}_{i}, {y}^{t}_{i})\}^{n_t}_{i=1}$ of $n_t$ training samples, with $\mathbf{x}^t_i \in \mathcal{X}_t,~ y^t_i \in \mathcal{L}_t$ and $\mathcal{L}_t = \{ l^t_i \}^{L_t}_{i=1}$ is the target label set.
In Inductive Transfer Learning we are also given a larger source set $\mathcal{S} = \{ (\mathbf{x}^{s}_i, {y}^{s}_i) \}^{n_s}_{i=1}$ of $n_s$ samples, with $ \mathbf{x}^s_i \in \mathcal{X}_s,~ y^s_i \in \mathcal{L}_s$ and $\mathcal{L}_s = \{ l^s_i \}^{L_s}_{i=1}$ is the source label set. (No specific assumption is made for $\mathcal{X}_t$ and $\mathcal{X}_s$, meaning they can either be equal or not equal.)  
We assume that $\mathcal{L}_s \neq \mathcal{L}_t$, this means that the target samples and source samples are from different concept categories. We also assume that a cost metric 
$c(\cdot,\cdot) : \mathcal{L}_s \times \mathcal{L}_t \to \mathbb{R}$ could be found, which indicates the relationships between each pair of source category and target category (We will show in section \ref{subsec:cost-metric} two examples on defining this cost metric). 

\begin{figure}[!htb]
  \centering
  \includegraphics[width=1\linewidth]{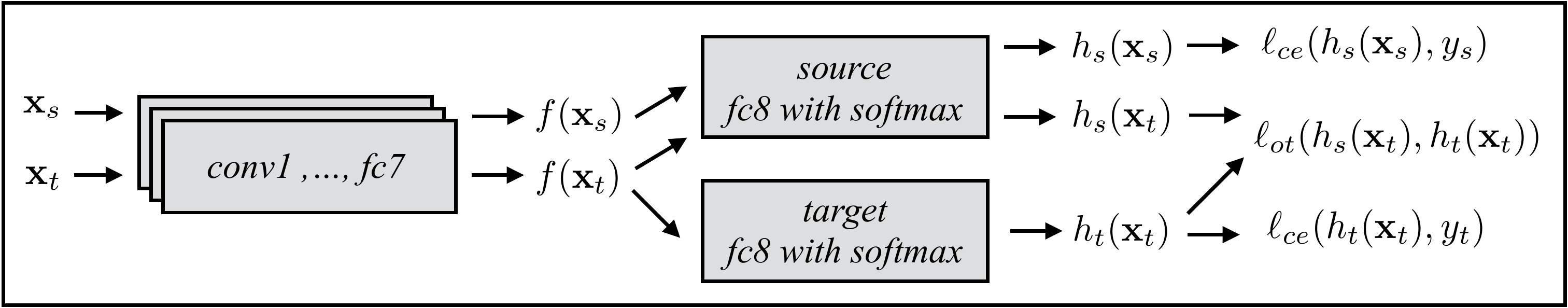}
  \caption{The structure and data flow of a Joint Transfer Learning Network based on Alexnet}
  \label{fig:JTLN_structure}
\end{figure}

We build the Joint Transfer Learning Network upon common Deep Neural Networks (\emph{e.g.} Alexnet for image classification), an illustration of a JTLN built upon Alexnet can be found in Figure \ref{fig:JTLN_structure}. In JTLN the feature extraction layers are shared by source data and target data and give $f(\mathbf{x}_i)$ as the feature vector for input sample $\mathbf{x}_i$. Following are two fully-connected layers with different output dimensions, which are considered as the source classifier and target classifier. The output of the source classifier is noted as: $h_s(\mathbf{x}_i) = a(\mathbf{W}_s \cdot f(\mathbf{x}_i) + \mathbf{b}_s)$, where $a(\cdot)$ is the softmax activation function, $\mathbf{W}_s$ and $\mathbf{b}_s$ are layer weight and bias for source classifier. The output of the target classifier is noted similarly: $h_t(\mathbf{x}_i) = a(\mathbf{W}_t \cdot f(\mathbf{x}_i) + \mathbf{b}_t)$, with $\mathbf{W}_t$ and $\mathbf{b}_t$ the layer weight and bias for target classifier. Two cross-entropy losses are added for joint learning with source data and target data. The source cross-entropy loss term is defined as: 
\begin{equation}
\label{eq:loss_src}
\frac{1}{n_s} \sum^{n_s}_{i=1} \ell_{ce}(h_s(\mathbf{x}^{s}_{i}),  {y}^{s}_{i})
\end{equation}
where $\ell_{ce}(\cdot,\cdot)$ is the cross-entropy loss function. The target cross-entropy loss term is defined similarly:
\begin{equation}
\label{eq:loss_tar}
\frac{1}{n_t} \sum^{n_t}_{i=1} \ell_{ce}(h_t(\mathbf{x}^{t}_{i}), {y}^{t}_{i})
\end{equation}

To express our prior knowledge about the relatedness between source and target tasks, we  propose to add a third Optimal Transport loss term for target data to restrict the distance between source classifier output and target classifier output, the OT loss term is noted as:
\begin{equation}
\label{eq:loss_w}
\frac{1}{n_t} \sum^{n_t}_{i=1} \ell_{ot}(h_s(\mathbf{x}^{t}_{i}),h_t(\mathbf{x}^{t}_{i}))
\end{equation}
where $\ell_{ot}(\cdot,\cdot)$ is the OT loss which will be defined in  section \ref{subsec:OTloss}. 

Therefore training with  JTLN is a problem of minimizing the empirical risk which is a combination of the three loss terms shown above:
\begin{equation}
\label{eq:JTLN_ERM}
\min_{\Theta} \frac{1}{n_t} \sum^{n_t}_{i=1} \ell_{ce}(h_t(\mathbf{x}^{t}_{i}), {y}^{t}_{i}) + \frac{\lambda_s}{n_s} \sum^{n_s}_{i=1} \ell_{ce}(h_s(\mathbf{x}^{s}_{i}), {y}^{s}_{i}) + \frac{\lambda_{ot}}{n_t} \sum^{n_t}_{i=1} \ell_{ot}(h_s(\mathbf{x}^{t}_{i}),h_t(\mathbf{x}^{t}_{i}))
\end{equation}

where $\Theta$ denote the set of all parameters in JTLN
, $\lambda_s$ is the loss weight for source cross-entropy loss and $\lambda_{ot}$ is the loss weight for OT loss.

\subsection{Optimal Transport Loss}
\label{subsec:OTloss}


In this paper we consider the discrete optimal transport problem. 
As the output of the source classifier (\emph{i.e.} $h_s(\mathbf{x}_i)$) and that of the target classifier (\emph{i.e.} $h_t(\mathbf{x}_i)$) are outputs of softmax activation, meaning that $\sum^{L_s}_{j=1} (h_s(\mathbf{x}_i))_j = 1$ and $\sum^{L_t}_{j=1} (h_t(\mathbf{x}_i))_j = 1$. We can therefore consider them as two probability measures over their corresponding label space. We define:
\begin{equation}
\label{eq:define_mu_nu}
\begin{split}
\mu = h_s(\mathbf{x}_i) \in \mathcal{Y}_s \\
\nu = h_t(\mathbf{x}_i) \in \mathcal{Y}_t
\end{split}
\end{equation}

where $\mathcal{Y}_s=\mathbb{R}^{L_s}_{+}$ is the space of measures over the source label set $\mathcal{L}_s$ and $\mathcal{Y}_t = \mathbb{R}^{L_t}_{+}$ is the space of measures over the target label set $\mathcal{L}_t$. Note that $\mu$ and $\nu$ defined here are discrete probability measures, \emph{i.e.} histograms in the probability simplex $\Delta^{\mathcal{L}_s}$ and $\Delta^{\mathcal{L}_t}$.

Given a cost metric $c(\cdot,\cdot) : \mathcal{L}_s \times \mathcal{L}_t \to \mathbb{R}$, the optimal transport problem aims at finding the optimal transportation plan $\gamma_0$ which minimizes the cost to transport the mass in probability measure $\mu$ to match that in $\nu$.
The Kantorovich formulation \cite{kantorovich2006translocation} of this discrete optimal transport problem can be defined as follows:

\begin{equation}
\label{eq:ot_exact}
\begin{split}
&\gamma_0 =  \argmin_{\gamma \in \Pi(\mu,\nu)} \langle \gamma, \mathbf{C} \rangle_F \\
\Pi(\mu,\nu) = \{ \gamma \in &(\mathbb{R}_+)^{L_s \times L_t} ~\vert~ \gamma \mathbf{1}_{L_t} = \mu,~ \gamma^\top \mathbf{1}_{L_s} = \nu \}
\end{split}
\end{equation}

where $\langle \cdot, \cdot \rangle_F$ is the Frobenius dot product. $\mathbf{C}$ is the cost matrix with $\mathbf{C}_{i,j} = c(l^s_i,l^t_j)$, the cost term $\mathbf{C}_{i,j}$, which can  be interpreted as the cost to move a probability mass from $l^s_i$ to $l^t_j$ (In our case we can think it as the cost to transfer the prediction on source category $l^s_i$ to the prediction on target category $l^t_j$). $\Pi(\mu,\nu)$ is the set of all valid transportation plans, \emph{i.e.} the set of joint probability measures on $\mathcal{L}_s \times \mathcal{L}_t$ with $\mu$ and $\nu$ as marginals. $\mathbf{1}_d$ is a $d$-dimensional vector of ones.

If $\mathcal{L}_s = \mathcal{L}_t$, the wasserstein distance can be defined and in \cite{frogner2015learning} the authors use the wasserstein distance formulation as a loss function for their multi-label prediction problem. In our case, we have assumed $\mathcal{L}_s \neq \mathcal{L}_t$, we therefore define our loss function in a similar way directly based on the Optimal Transport formulation:
\begin{defn}
\label{def:otloss}
(Optimal Transport Loss) ~\emph{For any source classifier $h_s: \mathcal{X} \to \Delta^{\mathcal{L}_s}$, and any target classifier $h_t: \mathcal{X} \to \Delta^{\mathcal{L}_t}$, given input $\mathbf{x} \in \mathcal{X}$, and a cost metric $c(\cdot,\cdot) : \mathcal{L}_s \times \mathcal{L}_t \to \mathbb{R}$, the Optimal Transport Loss is defined as:
\begin{equation}
\label{eq:otloss}
\ell_{ot}(h_s(\mathbf{x}),h_t(\mathbf{x})) \triangleq \inf_{\gamma \in \Pi(h_s(\mathbf{x}), h_t(\mathbf{x}))} \langle \gamma, \mathbf{C} \rangle_F
\end{equation}
where $\langle \cdot, \cdot \rangle_F$ is the Frobenius dot product. $\mathbf{C}$ is the cost matrix with $\mathbf{C}_{i,j} = c(l^s_i,l^t_j)$. 
$\Pi(h_s(\mathbf{x}),h_t(\mathbf{x}))$ is the set of valid transportation plans defined as:
\begin{equation}
\label{eq:otloss_pi}
\Pi(h_s(\mathbf{x}),h_t(\mathbf{x})) = \{ \gamma \in (\mathbb{R}_+)^{L_s \times L_t} ~\vert~ \gamma \mathbf{1}_{L_t} = h_s(\mathbf{x}),~ \gamma^\top \mathbf{1}_{L_s} = h_t(\mathbf{x}) \}
\end{equation}
where $\mathbf{1}_d$ is a $d$-dimensional vector of ones.}
\end{defn}

This Optimal Transport Loss  in Definition \ref{def:otloss} can therefore be calculated by solving the discrete optimal transport problem shown in \eqref{eq:ot_exact}. Problem in \eqref{eq:ot_exact} is a linear programming problem and can be solved with combinatorial algorithms, \emph{e.g.} the simplex methods and its network variants \cite{courty2016optimal}. However, the computational complexity is shown to be $O((L_s+L_t)L_s L_t log(L_s+L_t))$ at best \cite{ahuja1993network}. This limits the usage of this formulation for large scale dataset. 
Recently, Cuturi et al. \cite{cuturi2013sinkhorn} \cite{benamou2015iterative} proposed an entropy regularized optimal transport problem, which can be efficiently solved by iterative Bregman Projections. The discrete optimal transport problem with entropy regularization can be defined as:
\begin{equation}
\label{eq:ot_ereg}
\begin{split}
\gamma_0 &=  \argmin_{\gamma \in \Pi(\mu,\nu)} \langle \gamma, \mathbf{C} \rangle_F - \frac{1}{\lambda}H(\gamma)\\
\Pi(\mu,\nu) = \{ \gamma &\in (\mathbb{R}_+)^{L_s \times L_t} ~\vert~ \gamma \mathbf{1}_{L_t} = \mu,~ \gamma^\top \mathbf{1}_{L_s} = \nu \} \\
&H(\gamma) = \sum_{i,j} \gamma_{i,j} \log \gamma_{i,j}
\end{split}
\end{equation}

where $H(\gamma)$ is the entropy of $\gamma$ and $-\frac{1}{\lambda}$ is the regularization weight. 

This entropy regularization forces the solution of \eqref{eq:ot_ereg} to be smoother as $\frac{1}{\lambda}$ increase, \emph{i.e.}, as $\frac{1}{\lambda}$ increases, the sparsity of $\gamma^{\lambda}_0$ decreases. 
This non-sparsity of the solution helps to stabilize the computation by making the problem strongly convex with a unique solution. 
The advantage of this entropic regularized OT problem is that its solution is a diagonal scaling of $e^{-\lambda\mathbf{C}-1}$, where $e^{-\lambda\mathbf{C}-1}$ is the element-wise exponential matrix of $-\lambda \mathbf{C}-1$. The solution to this diagonal scaling problem can be found by iterative Bregman projections \cite{benamou2015iterative}.

In this work we calculate the approximation to the Optimal Transport Loss  in Definition \ref{def:otloss} by solving the entropic regularized optimal transport problem defined in \eqref{eq:ot_ereg} using iterative Bregman projections as shown in \cite{benamou2015iterative}. The computation of this approximate OT Loss is defined in Algorithm \ref{algo:aproxi_otloss}, where $./$ means element-wise division.

\begin{algorithm}
\renewcommand{\algorithmicrequire}{\textbf{Input:}}
\caption{Computation of the approximate OT Loss}
\label{algo:aproxi_otloss}
\begin{algorithmic}[1]
\REQUIRE 
$h_s(\mathbf{x}) \in \Delta^{\mathcal{L}_s}$, $h_t(\mathbf{x}) \in \Delta^{\mathcal{L}_t}$, $\lambda$, $\mathbf{C}$ \\
\STATE
Initialize: $\mathbf{u} = \mathbf{1}_{L_s} / L_s $, ~$\mathbf{v} = \mathbf{1}_{L_t} / L_t$, ~$\mathbf{K} = e^{-\lambda\mathbf{C}-1}$
\WHILE{$\mathbf{u}$ has not converged}
\STATE
$\mathbf{v} = h_t(\mathbf{x}) ./ (\mathbf{K}^{\top} \mathbf{u})$ 
\STATE
$\mathbf{u} = h_s(\mathbf{x}) ./ (\mathbf{K} \mathbf{v})$
\ENDWHILE
\STATE
$\ell_{ot}(h_s(\mathbf{x}),h_t(\mathbf{x})) = \langle diag(\mathbf{u}) \cdot \mathbf{K} \cdot diag(\mathbf{v}), \mathbf{C} \rangle_F$
\end{algorithmic}
\end{algorithm}

\subsection{Back-propagation with OT Loss}
\label{subsec:otloss-backprop}

The empirical risk minimization problem defined in Equation \ref{eq:JTLN_ERM} is normally solved by a gradient descent algorithm, therefore the gradient of each loss term with respect to their corresponding inputs should be expressed analytically for back-propagation. As in \cite{frogner2015learning} we define the Lagrange dual problem of LP problem \ref{eq:otloss} as :
\begin{equation}
\label{eq:ot_dual}
\begin{split}
\ell_{ot}(h_s(\mathbf{x}), h_t(\mathbf{x})) = \sup_{\alpha,\beta \in \mathcal{C}} {\alpha}^{\top} h_s(\mathbf{x}) + \beta^{\top} h_t(\mathbf{x}) \\
\mathcal{C} = \{ (\alpha,\beta) \in \mathbb{R}^{L_s \times L_t} :  \alpha_i + \beta_j \leq \mathbf{C}_{i,j}\}
\end{split}
\end{equation}

The $\mathbf{u}$ and $\mathbf{v}$ defined in Algorithm \ref{algo:aproxi_otloss} can be expressed as: $\mathbf{u} = e^{\lambda \alpha}$ and $\mathbf{v} = e^{\lambda \beta}$. As \ref{eq:otloss} is a linear program, at an optimum the values of the dual and the primal are equal, therefore the dual optimal $\alpha$ is a sub-gradient of the OT loss with respect to $h_s(\mathbf{x})$ and $\beta$ is a sub-gradient of the OT loss with respect to $h_t(\mathbf{x})$.

The gradient of OT loss with respect to its two arguments can therefore be expressed as follows and can be easily computed with the optimal scaling vectors $\mathbf{u}$ and $\mathbf{v}$ after matrix scaling with Algorithm \ref{algo:aproxi_otloss}:

\begin{equation}
\label{eq:otloss_gradients}
\begin{split}
\frac{\partial \ell_{ot}(h_s(\mathbf{x}),h_t(\mathbf{x}))}{\partial h_s(\mathbf{x})} = \alpha = \frac{\log \mathbf{u}}{\lambda} - \frac{\log \mathbf{u}^{\top} \mathbf{1}_{L_s}}{\lambda L_s} \mathbf{1}_{L_s} \\
\frac{\partial \ell_{ot}(h_s(\mathbf{x}),h_t(\mathbf{x}))}{\partial h_t(\mathbf{x})} = \beta = \frac{\log \mathbf{v}}{\lambda} - \frac{\log \mathbf{v}^{\top} \mathbf{1}_{L_t}}{\lambda L_t} \mathbf{1}_{L_t}
\end{split}
\end{equation}

Note that $\alpha$ and $\beta$ are defined up to a constant shift, \emph{i.e.} any upscaling of the vector $\mathbf{u}$ can be paired with a corresponding downscaling of the vector $\mathbf{v}$ (and vice versa) without altering the matrix $\gamma_0$, therefore the second terms in Equation \ref{eq:otloss_gradients} are added to ensure that $\alpha$ and $\beta$ are tangent to their corresponding simplex.

\subsection{Choosing the cost metric}
\label{subsec:cost-metric}
In Definition \ref{def:otloss}, the cost metric $c(\cdot,\cdot)$ can be interpreted as the cost to transfer the prediction on a source category to that of a target category. This cost metric embodies prior knowledge which describes the relatedness between each pair of source  and target categories. The choice of this cost metric is crucial in JTLN for having a better joint learning performance. 

A reasonable choice is to consider that the sample features in each category follow a probability distribution in a joint feature space, and to define this cost metric as the distance between two distributions. For example given a source category $l^s_i$ and a target category $l^t_j$, and a feature extractor $f(\mathbf{x})$ for sample $\mathbf{x}$. Suppose $\{f(\mathbf{x}^s) ~\vert~ \forall (\mathbf{x}^s,y^s),~ y^s = l^s_i \}$ follows the distribution $\mu_s$, and $\{f(\mathbf{x}^t) ~\vert~ \forall (\mathbf{x}^t,y^t),~ y^t = l^t_j \}$ follows the distribution $\mu_t$, our goal is to define a distance $d(\mu_s,\mu_t)$ between the two distributions as the cost metric for OT loss: $c(l^s_i,l^t_j) = d(\mu_s,\mu_t)$. To simplify the notations, in the following of this section we will use $\mathbf{x}^s$ and $\mathbf{x}^t$ instead of $f(\mathbf{x}^s)$ and $f(\mathbf{x}^t)$ to represent samples from the two distributions.

This definition implies that if the distribution of a target category and that of a source category lie close to each other in the feature space, their corresponding labels are more probably related and therefore cost less effort to transfer the prediction of one to that of the other. 

There are various ways to calculate the distance between two distributions. One way is to use the two-sample test with Multi-Kernel Maximum Mean Discrepancy (MK-MMD) as test statistics, which is successfully applied for solving domain adaptation problems \cite{long2015learning}. Another way is to use Optimal Transport and employ a basic distance metric (\emph{e.g.} Euclidean distance) as the cost metric. In the following we show details on how to apply these two methods as cost metrics for evaluating the distance between a given pair of source and target categories. 

\subsubsection{MK-MMD as cost metric}
\label{subsubsec:mkmmd}


Given samples from two distributions $\mu_s$ and $\mu_t$, a two-sample test determines whether to reject the null hypothesis $H_0: \mu_s=\mu_t$, based on the value of a test statistics measuring the distance between the samples. One choice of the test statistics is the maximum mean discrepancy (MMD), which is a distance between embeddings of the probability distributions in a reproducing kernel Hilbert space. Here we make use of the multi-kernel variant of MMD (MK-MMD) proposed in \cite{gretton2012optimal}, which maximizes the two-sample test power and minimizes the Type II error (\emph{i.e.}, the probability of wrongly accepting $H_0$ when $\mu_s \neq \mu_t$), given an upper bound on Type I error (\emph{i.e.}, the probability of wrongly rejecting $H_0$ when $\mu_s \neq \mu_t$), by leveraging different kernels for kernel embeddings.

Let $\mathcal{H}_k$ be a reproducing kernel Hilbert space (RKHS) endowed with a characteristic kernel $k$. The mean embedding of distribution $\mu_s$ in $\mathcal{H}_k$ is a unique element $\phi_k(\mu_s) \in \mathcal{H}_k$ such that $\mathbf{E}_{\mathbf{x} \sim \mu_s}g(\mathbf{x}) = \langle g(\mathbf{x}), \phi_k(\mu_s) \rangle_{\mathcal{H}_k}$ for all $g \in \mathcal{H}_k$. The MMD between probability distributions $\mu_s$ and $\mu_t$ is defined as the RKHS distance between the mean embeddings of $\mu_s$ and $\mu_t$. The squared formulation of MMD can be defined as:

\begin{equation}
\label{eq:mk-mmd}
d^2_k(\mu_s,\mu_t) = \parallel \phi_k(\mu_s) - \phi_k(\mu_t) \parallel^2_{\mathcal{H}_k} = \mathbf{E}_{\mathbf{x}^s{\mathbf{x}^s}'}k(\mathbf{x}^s,{\mathbf{x}^s}') + \mathbf{E}_{\mathbf{x}^t{\mathbf{x}^t}'}k(\mathbf{x}^t,{\mathbf{x}^t}') - 2\mathbf{E}_{\mathbf{x}^s \mathbf{x}^t}k(\mathbf{x}^s,\mathbf{x}^t)
\end{equation}

where $\mathbf{x}^s,{\mathbf{x}^s}' ~~\iidsim~~ \mu_s$ and $\mathbf{x}^t,{\mathbf{x}^t}' ~~\iidsim~~ \mu_t$. With $\phi_k$ an injective map, \emph{i.e.} $k$ is a characteristic kernel, the MMD is a metric on the space of Borel probability measures, \emph{i.e.} $d_k(\mu_s,\mu_t) = 0$ if and only if $\mu_s = \mu_t$. 
The characteristic kernel $k$ is defined as the convex combination of $m$ PSD (positive semi-definite) kernels $k_u$:
\begin{equation}
\label{eq:mkmmd-kernels}
\mathcal{K} \triangleq \{ k=\sum^m_{u=1}\beta_uk_u \vert \sum^m_{u=1}\beta_u=1, \beta_u \geqslant 0, \forall u \}
\end{equation}

where the constraints on coefficients $\{\beta_u\}$ are imposed to guarantee that the derived multi-kernel $k$ is characteristic. This multi-kernel $k$ can leverage different kernels to enhance the power of two-sample test.

For computation efficiency, we adopt the unbiased estimate of MK-MMD which can be computed with linear complexity:

\begin{equation}
\label{eq:mk-mmd-estim}
d^2_k(\mu_s,\mu_t) = \frac{2}{n}\sum^{n/2}_{i=1}g_k(\mathbf{z}_i)
\end{equation}

where $\mathbf{z}_i \triangleq (\mathbf{x}^s_{2i-1}, \mathbf{x}^s_{2i}, \mathbf{x}^t_{2i-1}, \mathbf{x}^t_{2i})$ is a random quad-tuple sampled from $\mu_s$ and $\mu_t$ 
, and we evaluate each quad-tuple with $g_k(\mathbf{z}_i) \triangleq k(\mathbf{x}^s_{2i-1}, \mathbf{x}^s_{2i}) + k(\mathbf{x}^t_{2i-1},\mathbf{x}^t_{2i}) - k(\mathbf{x}^s_{2i-1}, \mathbf{x}^t_{2i}) - k(\mathbf{x}^s_{2i},\mathbf{x}^t_{2i-1})$. 

\subsubsection{OT as cost metric}
\label{subsubsec:ot-costmetric}

Consider $\mu_s$ and $\mu_t$ as two empirical distributions defined by their corresponding discrete samples:

\begin{equation}
\label{eq:discrete_distributions}
\mu_s = \sum^{n_s}_{i=1}p^s_i \delta_{\mathbf{x}^s_i},~~ \mu_t = \sum^{n_t}_{i=1}p^t_i \delta_{\mathbf{x}^t_i}
\end{equation}

where $\delta_{\mathbf{x}_i}$ is the Dirac function at location $\mathbf{x}_i$. $p^s_i \in \Delta^{n_s}$ and $p^t_i \in \Delta^{n_t}$ are probability masses associated to the $i$-th sample. We can therefore define a discrete optimal transport problem with entropy regularization as in equation \eqref{eq:ot_ereg}:

\begin{equation}
\label{eq:otmetric_def}
\begin{split}
\gamma_0 &=  \argmin_{\gamma \in \Pi(\mu_s,\mu_t)} \langle \gamma, \mathbf{C} \rangle_F - \frac{1}{\lambda}H(\gamma)\\
\Pi(\mu_s,\mu_t) = \{ \gamma &\in (\mathbb{R}_+)^{n_s \times n_t} ~\vert~ \gamma \mathbf{1}_{n_t} = \mu_s,~ \gamma^\top \mathbf{1}_{n_s} = \mu_t \} \\
&H(\gamma) = \sum_{i,j} \gamma_{i,j} \log \gamma_{i,j}
\end{split}
\end{equation}

We define the cost metric in Equation \eqref{eq:otmetric_def} as squared Euclidean distance between two samples $\mathbf{C}_{i,j} = \parallel \mathbf{x}^s_i - \mathbf{x}^t_j \parallel^2_2$, and define the distance between the two distributions as $d(\mu_s,\mu_t) = \langle \gamma_0, \mathbf{C} \rangle_F$. This distance can be computed using the same matrix scaling procedure as in Algorithm \ref{algo:aproxi_otloss}.

\section{Experiments}
\label{sec:exps}
In this section we show the experiments of our proposed JTLN built upon Alexnet for Inductive Transfer Learning (ITL) with fine-grained image classification datasets.

We make use of the FGVC-Aircraft Dataset \cite{maji13fine-grained}. The dataset contains 10000 images of aircraft, with 100 images for each of 100 different aircraft model variants. These aircraft variants are from 30 different manufacturers. We build our Inductive Transfer Learning datasets upon this dataset by choosing the model variants from one manufacturer as the target domain categories, and consider the rest of the model variants as source domain categories. The images in source categories are all used for JTLN training, while the images in target categories are split into a subset for training and a subset for testing. We choose the two manufacturers with the most model variants to form two different ITL datasets, the characteristics of these ITL datasets are listed in Table \ref{tab:dataset_aircraft}, where the dataset name is indexed by the target manufacturer name. 


Note that all experiments perform fine-tuning based on an Alexnet model pre-trained with the ImageNet database. Further to the recommendations in \cite{yosinski2014transferable}, we fix the first three convolution layers since the features learned by these layers are general for different tasks, fine-tune the 4-th and 5-th convolutional layers with a small learning rate because the features learned in these two layers are less general, and fine-tune the 6-th and 7-th fully connected layers with a larger learning rate because the features learned in these layers are more task specific. The classification layers (fc8 with softmax) are trained from scratch.

We compare our proposed JTLN with three baseline methods:  the first one consists of fine-tuning only with  target training samples; the second one is the commonly adopted method, which first fine-tunes the pre-trained model with source samples, then continues fine-tuning with target training samples; the third baseline performs fine-tuning jointly with source samples and target training samples without applying the OT loss.

We also show the results of two variants of the proposed JTLN: 
(1) JTLN (fc7MKMMD) is JTLN using MK-MMD as cost metric as shown in section \ref{subsubsec:mkmmd}, and using the fc7-layer output of the Alexnet model pre-trained on ImageNet as features for the computation of the MK-MMD distances. 
(2) JTLN (fc7OT) is JTLN using OT as cost metric as shown in section \ref{subsubsec:ot-costmetric}, using the same Alexnet fc7-layer output as features. 
%
%



\begin{table*}[!htb] \footnotesize
\begin{minipage}{0.43\textwidth}
\centering
\caption{ITL Datasets with FGVC-Aircraft images}
\label{tab:dataset_aircraft}
\begin{tabular}{l|cc}
\Xhline{1.2pt}
\Gape[6pt][4pt]{\tabincell{l}{Dataset properties}} & Boeing & Airbus \\
\Xhline{1pt}
\Gape[4pt][4pt]{\tabincell{l}{N\textsuperscript{o} of target \\ categories}} & 22 & 13 \\
\hline
\Gape[4pt][4pt]{\tabincell{l}{N\textsuperscript{o} of target \\ training images}} & 1466 & 867 \\
\hline
\Gape[4pt][4pt]{\tabincell{l}{N\textsuperscript{o} of target \\ testing images}} & 734 & 433 \\
\hline
\Gape[4pt][4pt]{\tabincell{l}{N\textsuperscript{o} of source \\ categories}} & 78 & 87 \\
\hline
\Gape[4pt][4pt]{\tabincell{l}{N\textsuperscript{o} of source \\ images}} & 7800 & 8700 \\
\Xhline{1.2pt}
\end{tabular}
\end{minipage}
\hspace{18pt}
\begin{minipage}{0.5\textwidth}
\centering
\caption{Experimental Results on the ITL Datasets  (results  are multi-class classification accuracy)}
\label{tab:results_aircraft}
\begin{tabular}{l|cc}
\Xhline{1.2pt}
\Gape[8pt][6pt]{Methods} & Boeing & Airbus \\
\Xhline{1pt}
\Gape[4pt][4pt]{\tabincell{l}{Finetuning on target}} & 0.4796 & 0.4965 \\
\hline
\Gape[4pt][4pt]{\tabincell{l}{Consecutive finetuning\\on source+target}} & 0.5286 & 0.545 \\
\hline
\Gape[4pt][4pt]{\tabincell{l}{Joint finetuning\\on source+target}} & 0.5395 & 0.5497 \\
\hline
\Gape[4pt][4pt]{\tabincell{l}{JTLN (fc7MKMMD)}} & 0.5422 & \textbf{0.5982} \\
\hline
\Gape[4pt][4pt]{\tabincell{l}{JTLN (fc7OT)}} & \textbf{0.5436} & 0.5704 \\
\Xhline{1pt}
\end{tabular}
\end{minipage}
\end{table*}

The classification accuracies of these methods for the two ITL datasets are shown in table \ref{tab:results_aircraft}. We can see that with fc7MKMMD as cost metric,  JTLN for ITL-Airbus successfully improved the performance of joint fine-tuning by 5 points.  JTLNs (fc7MKMMD and fc7OT) on ITL-Boeing also improved in comparison with joint fine-tuning. However, the performance increase is not as high as that with the ITL-Airbus dataset.  We believe this  can  be partially explained by the fact  that  ITL-Boeing has less source categories and less source samples than  ITL-Airbus. 


\section{Conclusion and Future work}
\label{sec:conlu}
In this paper we have proposed a novel Joint Transfer Learning Network (JTLN) for Inductive Transfer Learning. By adding an Optimal Transport loss (OT loss) between the source and target classifier predictions during the joint fine-tuning process, the proposed JTLN can effectively learn useful knowledge for target tasks from source data. Another advantage of JTLN is the possibility of incorporating  prior knowledge about the relatedness between the target  and source categories by using different cost metric for OT loss. We show experimental results of JTLN with two different cost metrics in comparison with three baseline methods on two Inductive Transfer Learning datasets. The results verify the effectiveness of the proposed JTLN.

Future work includes further exploration of different cost metrics for OT loss. An interesting 
variant of JTLN could be to dynamically learn the cost matrix  along the fine-tuning process while using the current fine-tuned model as feature extractor. 


{\small
\bibliographystyle{unsrt}
\bibliography{JTLN}
}

\end{document}